\documentclass[conference]{IEEEtran}
\IEEEoverridecommandlockouts
\usepackage{cite}
\usepackage{amsmath,amssymb,amsfonts}
\usepackage{algorithmic}
\usepackage{graphicx}
\usepackage{textcomp}
\usepackage{xcolor}
\usepackage{hyperref}
\usepackage{subcaption}
\usepackage{orcidlink}
\def\BibTeX{{\rm B\kern-.05em{\sc i\kern-.025em b}\kern-.08em
    T\kern-.1667em\lower.7ex\hbox{E}\kern-.125emX}}
\newcommand{\SMART}{S.M.A.R.T. }

\begin{document}

\title{Large-scale End-of-Life Prediction of Hard Disks 
in Distributed Datacenters}

\author{\IEEEauthorblockN{Rohan Mohapatra}
\IEEEauthorblockA{\textit{Department of Computer Science} \\
\textit{San Jos\'e State University}\\
San Jose, CA, USA \\
rohan.mohapatra@sjsu.edu
}
\and
\IEEEauthorblockN{Saptarshi Sengupta}
\IEEEauthorblockA{\textit{Department of Computer Science} \\
\textit{San Jos\'e State University}\\
San Jose, CA, USA \\
saptarshi.sengupta@sjsu.edu}
}

\author{\IEEEauthorblockN{Rohan Mohapatra \orcidlink{0000-0003-1654-7994}\IEEEauthorrefmark{1} , Austin Coursey \orcidlink{0000-0003-1774-6442}\IEEEauthorrefmark{2} and
Saptarshi Sengupta \orcidlink{0000-0003-1114-343X}\IEEEauthorrefmark{1}}
\IEEEauthorblockA{\IEEEauthorrefmark{1}Department of Computer Science, San Jos\'e State University, San Jos\'e, CA, USA \\
\IEEEauthorrefmark{2}Department of Computer Science, Vanderbilt University, Nashville, TN, USA \\
Email: \IEEEauthorrefmark{1}rohan.mohapatra@sjsu.edu,
\IEEEauthorrefmark{2}austin.c.coursey@vanderbilt.edu, \IEEEauthorrefmark{1}saptarshi.sengupta@sjsu.edu}}

%%%%%%%%%%%%%%%%%%%%%%%%%%%%%%%%%%%%%%%%%%%%%%%%

\IEEEoverridecommandlockouts
% \IEEEpubid{\makebox[\columnwidth]{XXX-X-XXXX-XXXX-X/XX/\$XX.XX~\copyright2023 IEEE \hfill} \hspace{\columnsep}\makebox[\columnwidth]{ }}
\maketitle

\maketitle

\begin{abstract}
% Data centers process large amounts of data, the use of inexpensive hard-disks aided this. To mitigate the risk of failures, cloud storage providers prematurely replace hard-disks before they crash or fail. Sometimes, hard-disks fail, causing reliability issues that can hamper high-volume storage facilities. By understanding the remaining useful life (RUL) of Hard-disks, we can predict the failure day and replace it at the right time, ensuring maximum utilization whilst reducing operations costs. In this paper, we contrast the use of all health statistics against extracting useful attributes from severely skewed health statistics data. Past works suggest the use of LSTMs as an excellent approach to predict RUL. In this work, we also present an encoder-decoder LSTM model to predict RUL. The context gained from understanding health statistics sequences can predict an output sequence of the number of days remaining for a hard-disk. The model is trained and tested across all the years of data to build a robust model. We also tackle the problem of generalizability by using the trained LSTM on other hard-disk models.

On a daily basis, data centers process huge volumes of data backed by the proliferation of inexpensive hard disks. Data stored in these disks serve a range of critical functional needs from financial, and healthcare to aerospace. As such, premature disk failure and consequent loss of data can be catastrophic. To mitigate the risk of failures, cloud storage providers perform condition-based monitoring and replace hard disks before they fail. By estimating the remaining useful life of hard disk drives, one can predict the time-to-failure of a particular device and replace it at the right time, ensuring maximum utilization whilst reducing operational costs. In this work, large-scale predictive analyses are performed using severely skewed health statistics data by incorporating customized feature engineering and a suite of sequence learners. Past work suggests using LSTMs as an excellent approach to predicting remaining useful life. To this end, we present an encoder-decoder LSTM model where the context gained from understanding health statistics sequences aid in predicting an output sequence of the number of days remaining before a disk potentially fails. The models developed in this work are trained and tested across an exhaustive set of all of the 10 years of S.M.A.R.T. health data in circulation from Backblaze and on a wide variety of disk instances. It closes the knowledge gap on what full-scale training achieves on thousands of devices and advances the state-of-the-art by providing tangible metrics for evaluation and generalization for practitioners looking to extend their workflow to all years of health data in circulation across disk manufacturers. The encoder-decoder LSTM posted an RMSE of 0.83 during training and 0.86 during testing over the exhaustive 10 year data while being able to generalize competitively over other drives from the Seagate family.
\end{abstract}


\section{Introduction}
Distributed datacenters offer a broad range of services of critical importance to the public and private sectors alike. In order to support the voluminous day-to-day operations, the reliability of functional storage devices is of utmost importance. Data-center outages are plagued by Hard Disk Drive (HDD) failure events \cite{hdd-errors}. Accurate prediction of HDD life span allows timely mitigation measures that prevent data loss and improves data center reliability. Datacenters often rely on S.M.A.R.T. (Self-Monitoring Analysis and Reporting Technology) logs to understand how HDDs are performing. As seen by \cite{biderectionallstm}, Backblaze \cite{backblaze2022afr} reported an Annualized Failure Rate (AFR) of 1.37\%. The above AFR  means that, on average, 1.37\% of the hard drives in their datacenters fail each year. Predicting in advance when hard drives fail is beneficial because it allows for proactive maintenance and management and in turn can help reduce downtime, minimize the risk of data loss, and ultimately save time and money. Additionally, it can help identify patterns from failure logs, which can be used to inform future purchasing decisions and improve overall data storage strategies. This way, one is able to identify significant correlations and patterns from the physics of failure as well as produce helpful forecasts of impending faults. Such knowledge enables early intervention and helps keep downtime to a minimum.

% In this paper, ...

\textbf{Contributions:} The contributions of this work are significant in the following ways:

\begin{enumerate}
    \item \textit{\textbf{A generalizable encoder-decoder LSTM for RUL prediction:}} We propose a Long Short-Term Memory (LSTM) \cite{lstm} architecture based on an encoder-decoder scheme for the prediction task at hand. In contrast to traditional LSTM models that process input sequences one step at a time, the encoder-decoder LSTM processes the entire input sequence at once and generates the output sequence in a one-shot manner, thereby retaining its capability of learning complex temporal patterns over long windows. To the best of our knowledge, this type of network has not been implemented before for HDD RUL prediction using the comprehensive 10-year quarter-by-quarter health data from Backblaze.
    \item \textit{\textbf{Investigative analysis on the data:}} The Backblaze dataset is very large ( $\sim$ 35 GB) and varied, and as such, it is infeasible to process all inter-dependencies in it at once. In this context, we focused on analyzing Seagate models from 2013-2022. To analyze the data, a pipeline with PostgreSQL was proposed to preprocess and filter the Seagate models with significant failures. XGBoost \cite{xgboost} was employed for feature selection and interpolation was utilized to fill in the missing data. 
    \item \textit{\textbf{Generalizability of the network:}} Our networks are able to generalize well within a specific manufacturer's device family, raising the possibility of having multiple models working in tandem to estimate the underlying degradation of each manufacturer on an individual basis.

\end{enumerate}

The paper is organized in the following way. In Section \ref{section:lit}, we go over related work in the field. Section \ref{section:data} introduces a data pipeline, analysis of the failures of different Seagate models, and pre-processing techniques to improve the use of raw health statistics. Section \ref{section:lstm} talks about the rationale behind using an encoder-decoder LSTM and presents a hypothesis on the working of the proposed model. Section \ref{section:experiments} divulges into the different configurations and experimentation done with the dataset. In this section, we also analyze the generalizability of the proposed model and provide metrics to gain insights. Lastly, Section \ref{section:future} looks at potential avenues for future research setting this paper as a baseline.
% Firstly, it presents a review of the existing literature on the topic, synthesizing and evaluating previous studies and theories to provide a clear understanding of the research problem. Secondly, it introduces an approach with encoder-decoder LSTMs to predict end-of-life sequences of the number of days remaining before a disk fails. Thirdly, the paper presents a thorough investigation and experimentation on 10 years of data and insights on the generalizability of the proposed model on a number of models in the Seagate family. Finally, the paper directs future work on attention-based mechanisms, threat modeling to increase robustness, and an ensemble approach. 

\section{Related Work}
\label{section:lit}
S.M.A.R.T. is widely recognized across the industry as the standard monitoring and failure warning technology for hard disk drives \cite{SMART}. However, how many computer systems today enable or read the \SMART logs is unknown. This is not the case for datacenters that use threshold testing algorithms to detect drive failures \cite{thresholdtestingalgo}. Classical approaches include using an SVM to classify failures \cite{svm} and Bayesian-based prediction of failures \cite{bayesian}. To make forecasts, \cite{hmmhdd} observed S.M.A.R.T. characteristics as time series and use Hidden Markov Models (HMMs) and Hidden Semi-Markov Models (HSMMs) to construct models for "good" and "failed" drives. The increasing tendency in attribute values (or their rates of change) over time indicates that some attributes are about to fail. Of late, researchers have begun investigating the use of data-driven methods, specifically recurrent neural networks such as LSTM networks, to forecast hard drive failures based on a study of sensor data incorporating S.M.A.R.T characteristics. This serves as a prelude to the use of deep recurrent networks and advanced LSTM architectures to detect drives that may fail by predicting their RUL \cite{biderectionallstm, hddpaper4, hddpaper2, stackedlstm, BASAK2021101283}.

A fair amount of research has been done by using LSTMs to forecast hard drive failures. Lima et al.\cite{stackedlstm} used a stacked LSTM to predict the Remaining Useful Life (RUL) of HDDs. Such use is motivated by the network's ability to handle complicated temporal relationships in data. This is essential for forecasting hard drive failures because such failures are often a gradual process that can be affected by a variety of variables over time. Very recently, Coursey et al. \cite{biderectionallstm} explored the possibility of using Bidirectional LSTMs. Bi-LSTM is especially useful for describing temporal relationships in time-series data as these networks extend the conventional LSTM design by processing input sequences in both forward and backward directions, enabling them to record both past and future context in data. The temporal relationships between S.M.A.R.T. characteristics in the case of hard drive failure forecast can be complicated and non-linear, and it is essential to consider both past and future circumstances when making statements about an HDD's remaining useful life. The proposed model can successfully encapsulate these dependencies and make accurate forecasts.

\section{Understanding the Data}
\label{section:data}
The hard drive dataset is provided by Backblaze and it contains logs recorded daily at the datacenter level. The data is collected from 2013-2022 and contains information about the date, model, serial number, S.M.A.R.T. features, and if the hard drive has failed. In this paper, we consider Seagate models, pre-process and clean the data before we feed it to any network. On a daily basis, any hard drive logs the following metrics: (a) the timestamp of the log, (b) the serial number and model of the hard drive, (c) failure label: 1 if the HDD has failed and 0 otherwise, (d) S.M.A.R.T. features: All S.M.A.R.T. features are included in the dataset. Some of them are missing or have zero values.

Seagate is one of the major manufacturers of hard drives and has been collecting data on hard drive failures for many years. This means that there is a large amount of data available on Seagate hard drives for researchers to use in studies of reliability and predict remaining useful life. In addition, Seagate is well-known to have built a reputation for producing hard drives with consistent performance characteristics, which in turn can make the data collected from Seagate drive logs quite reliable for use in RUL prediction studies. Even if we pivot to another manufacturer, Backblaze over the years has collected more \SMART logs for Seagate than for any other manufacturer. Because of this, our analysis on RUL predictions is focused on Seagate hard drives.

\subsection{Data Pipeline: Processing Data}
\begin{figure}[h!]
\centering
\includegraphics[width=0.75\columnwidth]{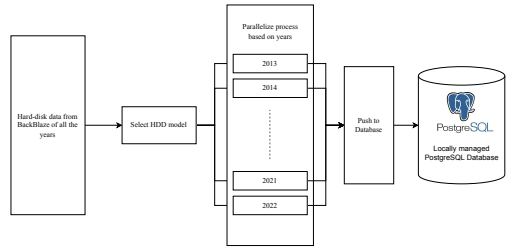}
\caption{Data pipeline to process raw sensor data}
\label{fig:datapipeline}
\end{figure}

The comprehensive Backblaze dataset is very large ($\sim$ 35 GB in size) and reading it to memory on standard workstations is not a viable option. An easy way to extract data based on specific years for a given hard-disk model warranted use of a database. We processed the data for each model into a separate logical layer using the pipeline shown in Fig. \ref{fig:datapipeline}. An HDD model was selected which was used to filter data out. This filtered data was then parallelized based on year of reporting since Backblaze reports health statistics every year. It was then pushed to a locally managed PostgreSQL server.

\subsection{Extracting HDD Models with Failures}
The next task in understanding the data is to figure out the HDD models that fail more frequently. Using a modified data pipeline from Fig, \ref{fig:datapipeline}, we queried for the number of unique failures of all Seagate models. Table \ref{table:failures} shows the Seagate models and the number of unique failures from 2013-2022.

\begin{table}[htbp]
\begin{center}
\caption{Unique failures of Seagate models (2013 to 2022)}
\begin{tabular}{|c|c|}
\hline
\textbf{Seagate model}             & \textbf{Unique failures} \\ \hline \hline
ST4000DM000                            & 4934                               \\ \hline
ST12000NM0007                          & 2010                               \\ \hline
ST3000DM001                            & 1708                               \\ \hline
ST8000NM0055                           & 1101                               \\ \hline
ST8000DM002                            & 731                                \\ \hline
ST12000NM0008                          & 679                                \\ \hline
ST31500541AS                           & 397                                \\ \hline
ST31500341AS                           & 216                                \\ \hline
...                                    & ...                                \\ \hline
\end{tabular}
\label{table:failures}  
\end{center}
\end{table}

\subsection{Feature Selection using XGBoost}
The dataset contains $\sim$30 S.M.A.R.T features. As part of our pre-processing and visualization, we realized that working with all the features could be a viable approach. But as we looked through different HDD models, we found that there are very few overlapping features. Another thing that we realized was that many S.M.A.R.T. features were either zeros or did not vary with time. This led us to believe that a feature selection method could isolate the important features. We subsequently used XGBoost  \cite{xgboost} to select these important features. However, we do present a comparison of training with all features versus only the important ones for the sake of completeness.

\begin{figure}[h!]
\centering
\includegraphics[width=0.7\columnwidth]{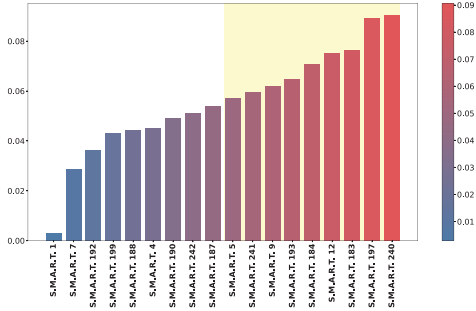}
\caption{Features sorted by importance weights}
\label{fig:impfeat}
\end{figure}

% XGBoost, or Extreme Gradient Boosting, is a powerful algorithm that is often used for feature extraction in machine learning. It is particularly good for this task because it can handle high-dimensional feature spaces as in the case of the hard disk dataset and is able to automatically select the most important features. XGBoost works by iteratively adding decision trees to a model, with each new tree trying to correct the errors made by the previous trees. The algorithm places more emphasis on misclassified data points, which makes it robust against outliers and noise. Additionally, XGBoost allows for the computation of feature importance scores, which can be used to rank features and select the most relevant ones for a particular task. It uses a built-in feature importance metric that allows it to rank the input features based on how much they contribute to the model's performance which in this case, by intuition, provides relevant \SMART features.

Some of the important features that we discovered are tabulated in Table \ref{table:smart}. While there are numerous \SMART features that provide information about a hard drive's health, only a subset of them is commonly associated with potential drive failure as shown in Fig. \ref{fig:impfeat}. These specific \SMART features provide insight into aspects of a hard drive's health, such as the frequency of raw read errors, reallocated sectors, start-stop cycles, spin-up time, power-on hours, and temperature fluctuations, among others. These factors can all contribute to the gradual degradation of a hard drive's mechanical components, ultimately leading to drive failure. For instance, an excessively high raw read error rate, which indicates errors encountered while reading data from a hard drive's platters, can be an early warning sign of drive failure. Similarly, an increased number of reallocated sectors suggests that a hard drive is struggling to maintain the integrity of stored data, which can also lead to a hard drive failure. 
% \begin{enumerate}
%     \item S.M.A.R.T 1 
%     \item S.M.A.R.T 4
%     \item S.M.A.R.T 5 
%     \item S.M.A.R.T 7 
%     \item S.M.A.R.T 9
%     \item S.M.A.R.T 188 
%     \item S.M.A.R.T 190
%     \item S.M.A.R.T 241
%     \item S.M.A.R.T 242
% \end{enumerate}

\begin{table*}[h]
\centering
\caption{Selected \SMART features and their context related to hard drives \cite{backblaze2014}}
\scalebox{0.7}{\begin{tabular}{|l|p{15cm}|}
\hline
\textbf{\SMART Feature} & \textbf{Meaning} \\
\hline
\hline
\SMART 1 & Raw Read Error Rate: The total number of errors the hard drive encounters when reading data from a disk. \\
\hline
\SMART 4 & Start/Stop Count: The total number of times the hard drive is powered on and off. \\
\hline
\SMART 5 & Reallocated Sector Count: The total number of sectors that have been remapped due to read errors. \\
\hline
\SMART 7 & Seek Error Rate: The total number of errors when positioning the read/write head on the disk sector. \\
\hline
\SMART 9 & Power-On Hours: The total number of hours the hard drive has been powered on. \\
\hline
\SMART 188 & Command Timeout: The total number of times that the drive did not respond to a command sent by the computer's OS. \\
\hline
\SMART 190 & Airflow Temperature Celsius: The temperature of the hard drive's airflow, measured in Celsius. \\
\hline
\SMART 192 & Power-Off Retract Count: The total number of times the read/write head is retracted due to power loss or other reasons. \\
\hline
\SMART 193 & Load/Unload Cycle Count: The total number of times the head is loaded and unloaded. \\
\hline
\SMART 194 & Temperature Celsius: The temperature of the hard drive, measured in Celsius. \\
\hline
\SMART 197 & Current Pending Sector Count: The total number of unstable sectors that the hard drive is attempting to read. \\
\hline
\SMART 198 & Offline Uncorrectable Sector Count: The total number of sectors that cannot be corrected using hardware error correction. \\
\hline
\SMART 199 & UDMA CRC Error Count: The total number of errors when data is transferred between the hard drive and the computer's memory. \\
\hline
\SMART 241 & Total LBAs Written: The total number of sectors written to the hard drive. \\
\hline
\SMART 242 & Total LBAs Read: The total number of sectors read from the hard drive. \\
\hline
\end{tabular}}
\label{table:smart}
\end{table*}

\begin{figure}[h!]
\includegraphics[width=\columnwidth]{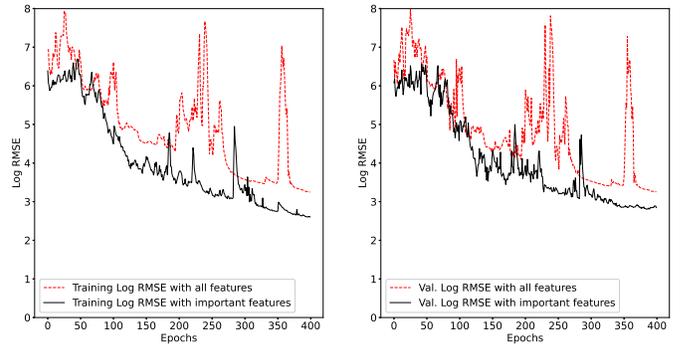}
\caption{Training and validation RMSE using all features vs important features discovered by XGBoost}
\label{fig:fullvimp}
\end{figure}

Fig. \ref{fig:fullvimp} demonstrates that using important features translates to a lower train and validation RMSE (2022 logs)

\subsection{Missing Data}
\begin{figure}[htbp]

\begin{subfigure}[b]{0.5\textwidth}
\centering
        \includegraphics[width=0.65\columnwidth]{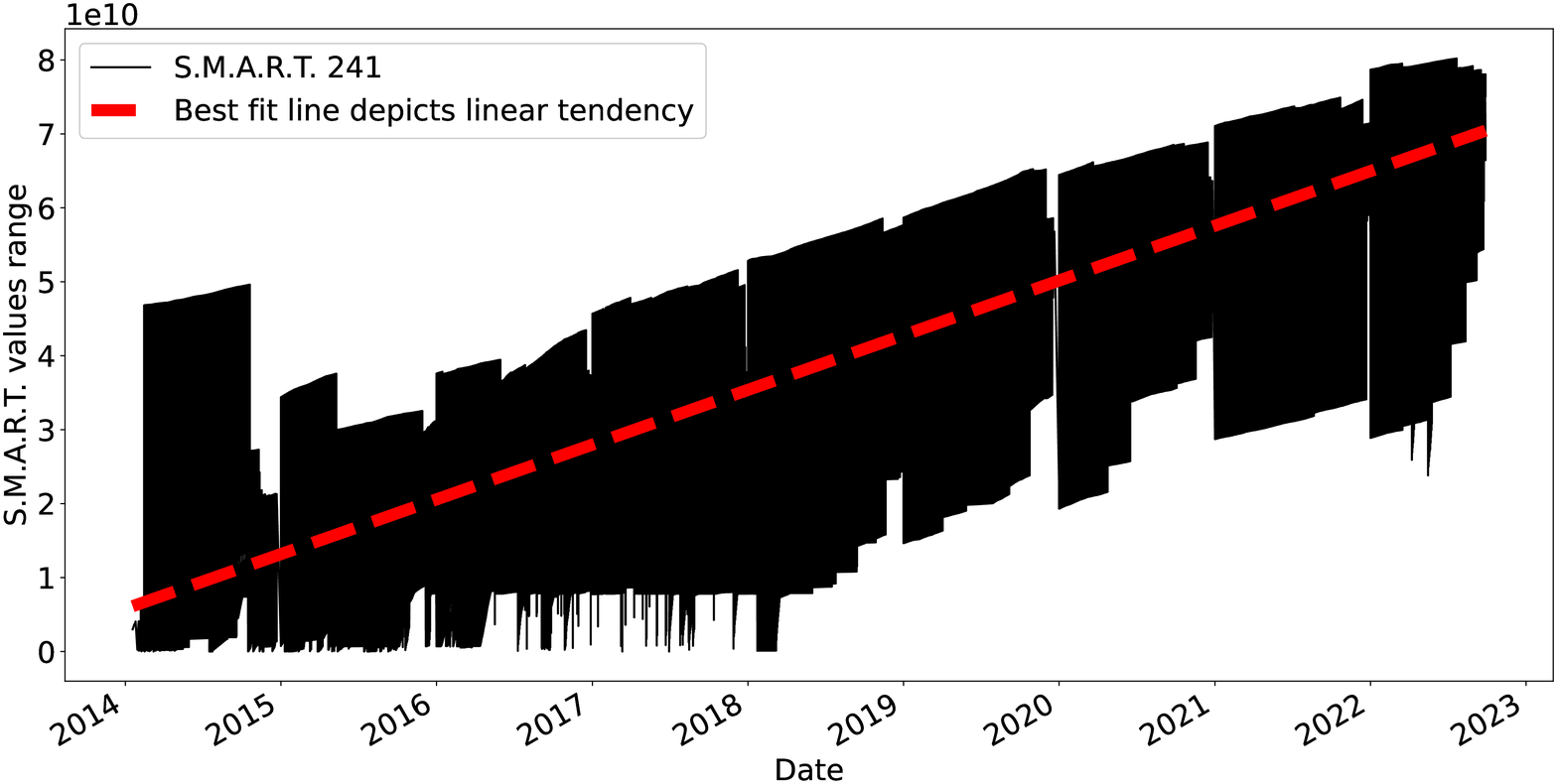}
   \caption{}
   
   \label{fig:lineartendency} 
\end{subfigure}
\begin{subfigure}[b]{0.5\textwidth}
\centering
   \includegraphics[width=0.55\columnwidth]{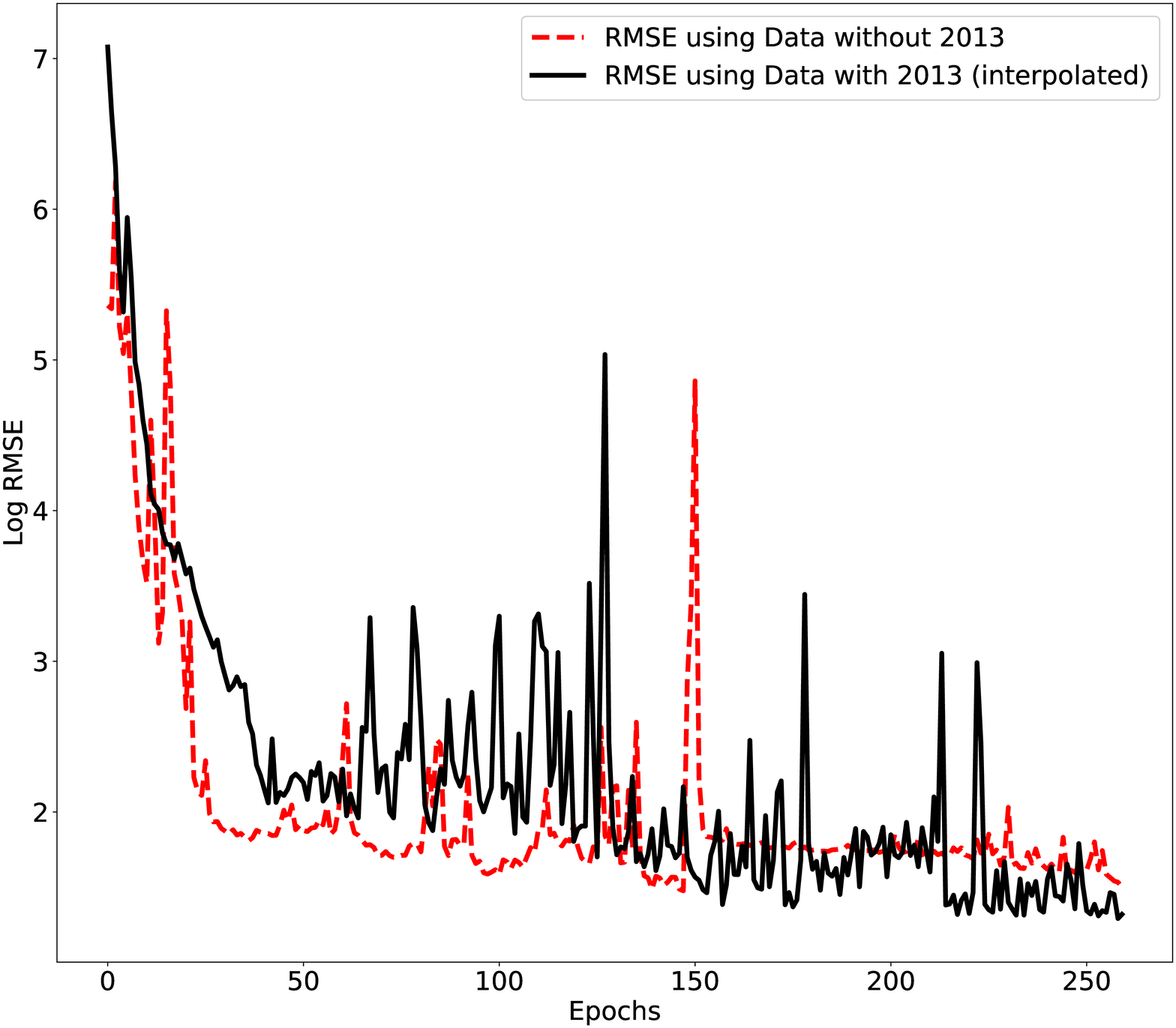}
   \caption{}
   \label{fig:interpolate}
\end{subfigure}
\caption{(a) An example of \SMART 241 feature, this figure demonstrates the linear tendency (b) RMSE trend when interpolation is used}
\end{figure}

A known issue that has been carefully ignored with the Backblaze dataset is the missing S.M.A.R.T. parameters before February 2014. Santo et al. \cite{hddpaper4} conclusively indicate that 70 \% of the S.M.A.R.T. parameters had not been collected for the Seagate model in focus during this time. As we try to reason whether the data before this time is meaningful or not, we train on the data with and without the missing data (gaps filled by interpolation) to come to a conclusion. Since the time series has a linear tendency (see Fig. \ref{fig:lineartendency}), we use linear interpolation to replace the empty gap. As noticed from Fig. \ref{fig:interpolate}, the interpolation indeed works and gives a substantial extra year to extract meaningful predictions. Lower RMSE is also an indication that data points lying between 2013 and February 2014 are significant and cannot be simply ignored.

% When there are gaps in the data, a baseline method prior to statistical analyses is to use interpolation to fill those gaps. Interpolation can be used in time series analysis to approximate missing values at a particular time point using the values of nearby time points. Linear interpolation is the most basic technique for estimating missing values by making a straight line between two known values on either side of the missing value.

% \begin{figure}[htbp]
% \includegraphics[width=\columnwidth]{interpolate.eps}
% \caption{RMSE trend when interpolation is used to fill the missing gaps}
% \label{fig:interpolate}
% \end{figure}

\subsection{Data Scaling}
Historically, data scaling is performed using a minimum maximum scaler that clamps the data between 0 and 1 \cite{biderectionallstm}. This would work best and provide a lower RMSE value. But as we observed, the data is not exactly between a small range which can translate to being normalized between 0-1. Such a scaling would most likely lead to information loss. On analyzing the important features, we observed that the \SMART features range from 0 to the order of $10^{14}$. Such a huge difference between the maximum and minimum values cannot be scaled to [0, 1] without potential information loss. Adopting the normalization technique proposed by Backblaze \cite{beach_2021}, we normalized and scaled the data between [0, 255] and we believe this provides more information for the model to learn and predict without bias.

\begin{figure*}[h!]
 \centering
\includegraphics[width=0.65\textwidth]{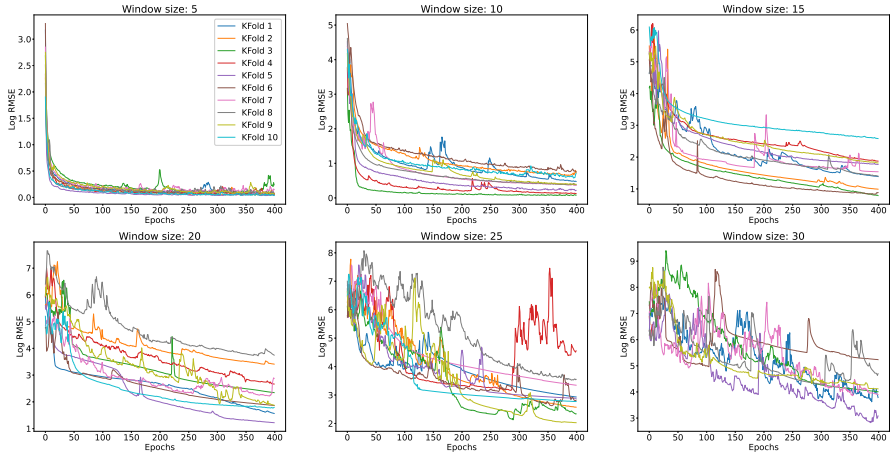}
\caption{Training RMSE with varying window sizes}
\label{fig:window}
\end{figure*}

\section{The Machine Learning Approach}
\label{section:lstm}
The remaining useful life prediction extends to a time series forecasting problem. In order to make forecasts and guide strategic decision-making, time series forecasting involves studying time series data using statistics and modeling. Of late, machine learning approaches using deep neural networks have been extensively used for time series forecasting \cite{HEWAMALAGE2021388}. Through time, LSTMs have gained popularity for such problems. LSTM is a type of recurrent neural network popularly used in deep learning because it can learn sequence dependencies efficiently and successfully. 

An LSTM block contains 3 gates, \textit{input gate} ($i_t$), \textit{forget gate} ($f_t$) and \textit{output gate} ($o_t$). An LSTM also contains a \textit{memory cell} ($C_t$) introduced as a measure to counter the problem of vanishing/exploding gradient found in a simple RNN cell.  

\begin{equation}
f_t = \sigma_g(W_f h_{t-1} + U_f x_t + b_f) 
\end{equation}
\vspace{-6mm}
\begin{equation}
i_t = \sigma_g(W_i h_{t-1} + U_i x_t + b_i) 
\end{equation}
\vspace{-6mm}
\begin{equation}
\tilde{C}_t = \tanh(W_C h_{t-1} + U_C x_t + b_C) 
\end{equation}
\vspace{-6mm}
\begin{equation}
C_t = f_t \odot C_{t-1} + i_t \odot \tilde{C}t 
\end{equation}
\vspace{-6mm}
\begin{equation}
o_t = \sigma_g(W_o h_{t-1} + U_o x_t + b_o) 
\end{equation}
\vspace{-6mm}
\begin{equation}
h_t = o_t \odot \tanh(C_t)
\end{equation}

where:
\begin{enumerate}
\item $x_t$ is the input at time $t$
\item $h_{t-1}$ is the hidden state at time $t-1$
\item $f_t$ is the forget gate output at time $t$
\item $i_t$ is the input gate output at time $t$
\item $\tilde{C}_t$ is the candidate cell state at time $t$
\item $C_t$ is the cell state at time $t$
\item $o_t$ is the output gate output at time $t$
\item $h_t$ is the hidden state output at time $t$
\end{enumerate}

We employ an advanced LSTM architecture called Encoder-Decoder LSTM \cite{NIPS2014_a14ac55a} or Seq2Seq LSTM for RUL prediction.
\subsection{Encoder-Decoder LSTM (Sequence to Sequence LSTM)}
% LSTM networks are recurrent neural networks that are capable of recording temporal relationships in time series data. They are made up of cells and hidden states that enable the accounting of long and short-term memory impacts. A big improvement in language-to-language translation was achieved by Encoder-Decoder LSTM . 
The Encoder-Decoder (ED) structure employs two LSTM networks in the encoder and decoder portions. This framework enables the model to manage a variety of input and output time steps, which can be helpful for ahead-of-time forecasting. In this model, the encoder layer only emits the last cell's hidden state, which is then used as input for each LSTM cell in the decoder layer, allowing information from the input series to be collected at each time step. In comparison to the regular LSTM, structure improves long-term dependencies for extended time step forecasts. This research used five extra dense layers after the LSTM decoder layer to improve the decoding of sequence output at each time increment. 

\begin{figure}[h!]
\centering
\includegraphics[width=0.85\columnwidth]{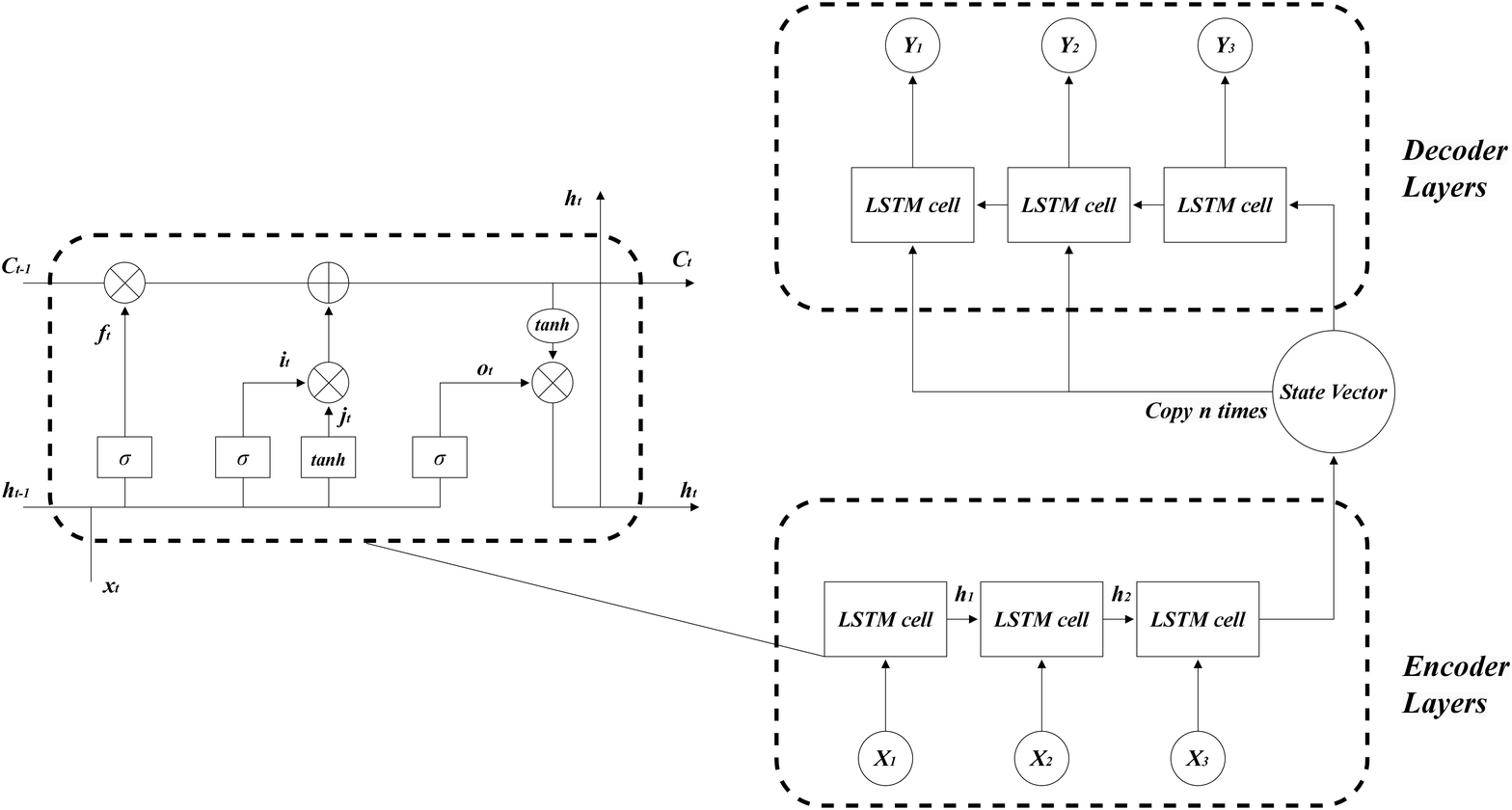}
\caption{Architecture of an encoder-decoder LSTM}
\label{fig:architecture}
\end{figure}
\begin{figure}[h!]
\centering
\includegraphics[width=0.65\columnwidth]{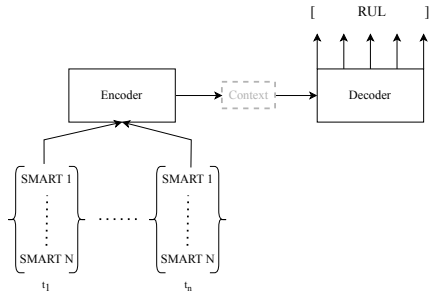}
\caption{Architecture with the customized inputs and outputs}
\label{fig:end-dec}
\end{figure}

Building upon the existing knowledge, one could leverage the latent space encoding of past observations of S.M.A.R.T. to incorporate more contextual information for improved RUL forecasting. This can be achieved by first encoding the S.M.A.R.T values from $t$ (time-steps) days into a latent space using an encoder, then utilizing the encoded past as a "context" in the decoder to improve RUL predictions. The decoder could be modified to take the encoded past as input in addition to the temporal data, thereby allowing for a richer representation of past patterns to inform future predictions. By incorporating this additional information, the model may be better equipped to handle more complex and nuanced patterns in the data, potentially leading to more accurate RUL forecasts. This also opens the avenue to look at attention mechanisms to further enhance the training and build upon this model. Fig. \ref{fig:end-dec} describes the idea and the approach that the rest of the paper takes while building the model.

% \begin{figure*}[htbp]
%  \centering
% \includegraphics[height=0.65\textwidth]{RUL_4.eps}
% \caption{Predicted vs. Expected RUL on a test set for the models mentioned in Table \ref{table:configurations} }
% \label{fig:Ng2}
% \end{figure*}

\begin{figure*}[htbp]
\centering
\begin{subfigure}[b]{\textwidth}
        \centering
        \includegraphics[width=0.75\columnwidth]{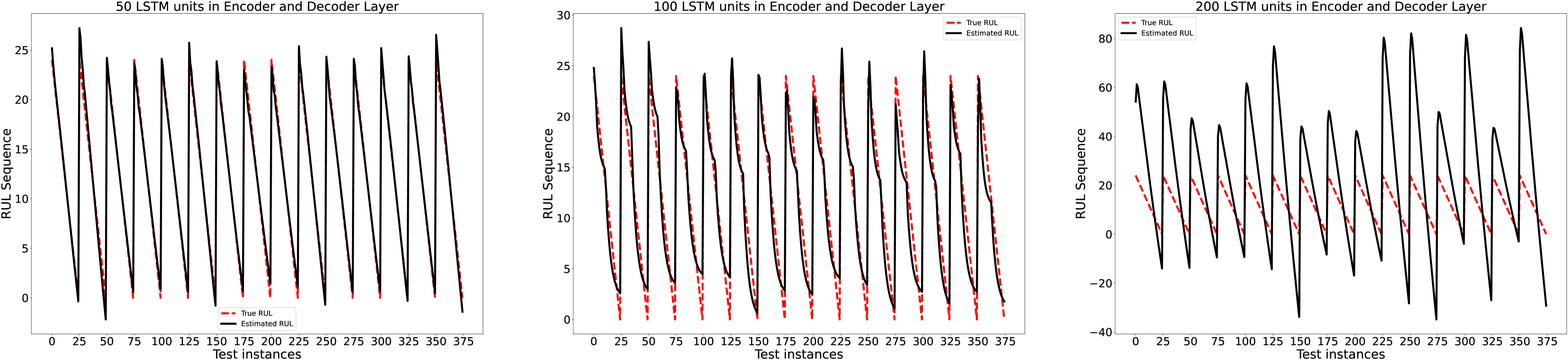}
   \caption{Experiments run with a single encoder-decoder with varying LSTM units}
\end{subfigure}
\\~\\
\begin{subfigure}[b]{\textwidth}
\centering
   \includegraphics[width=0.75\columnwidth]{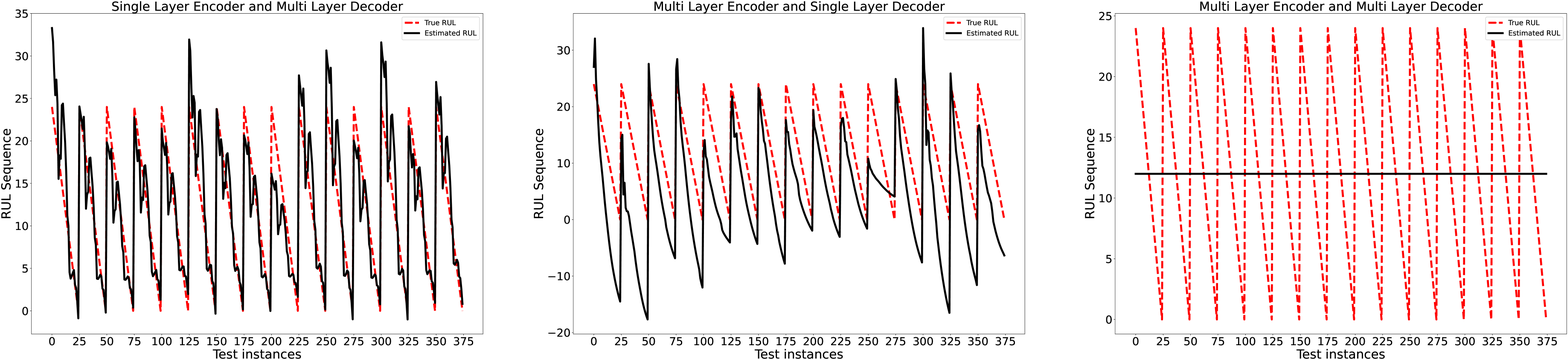}
   % [width=0.85\columnwidth]
   \caption{Experiments run with encoder-decoder LSTM with multiple layers}
\end{subfigure}
\caption{Predicted vs. expected RUL on a test set for the models mentioned in Table \ref{table:configurations} }
\label{fig:Ng2}
\end{figure*}

\begin{figure*}[htbp]
 \centering
\includegraphics[width=0.75\textwidth]{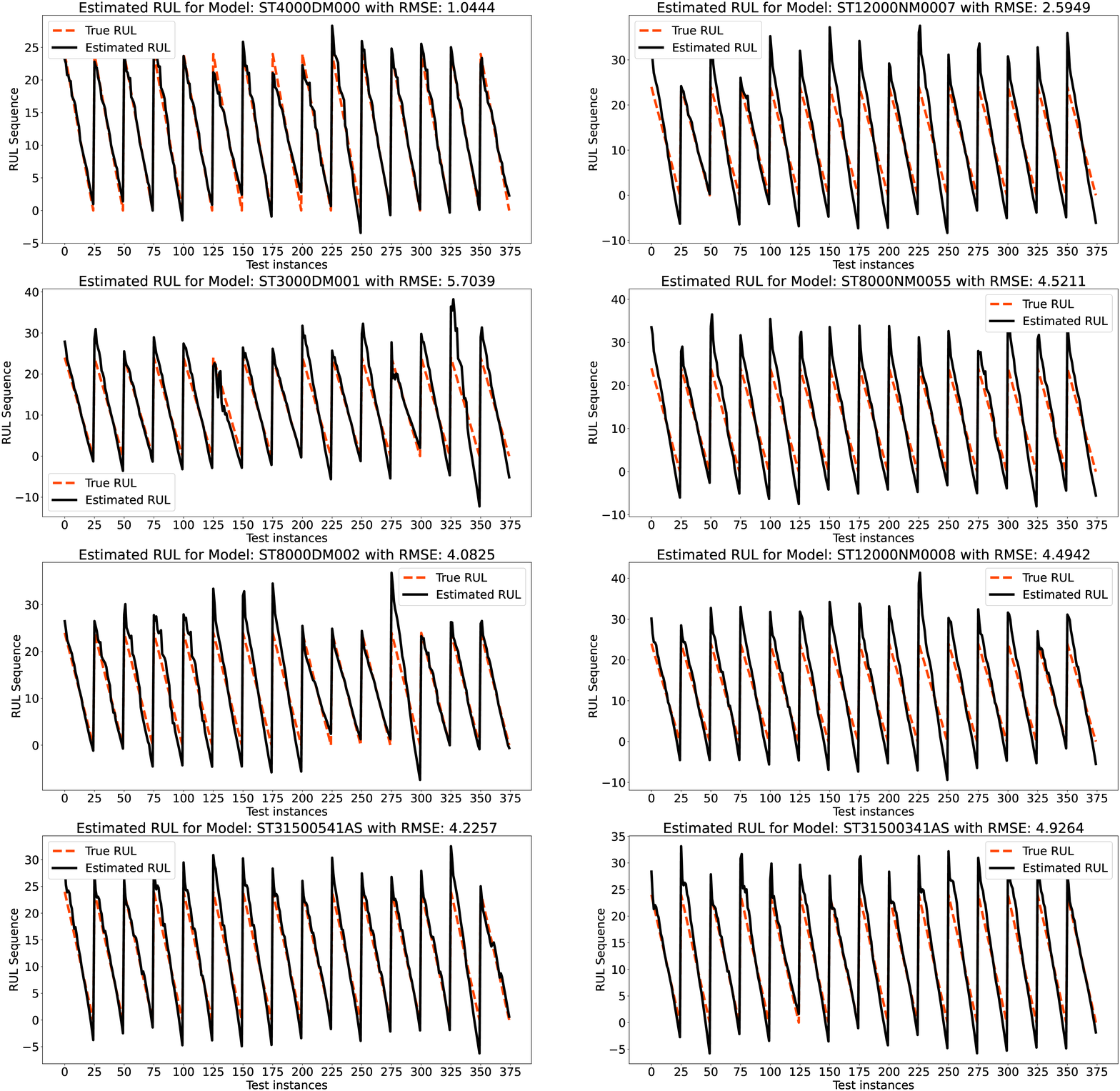}
\caption{Predicted vs expected RUL on different models of the Seagate family }
\label{fig:generalize}
\end{figure*}

\section{Experiments}
\label{section:experiments}
In the above section, we explain the data pre-processing stages and pipeline. Deep neural networks can be computationally expensive to train when the training data size is large. Careful selection of the batch size and techniques like early stopping \cite{Prechelt1996EarlySW} can help train LSTMs with little or no overfitting. We build multiple Encoder-Decoder LSTMs and analyze their impact on learning. In this section, we break it into different subsections where we first analyze thoroughly a subset of data and then extend it to the complete dataset. 

For the analysis of a subset of the data, we look at 2022 Q1 and Q2 data, with the model most vulnerable to failure (ST4000DM000). Typically, an LSTM model requires data to be in the shape of [\textit{samples} x \textit{timesteps} x \textit{features}]. \cite{biderectionallstm} describes in brief what each of these parameters means.

\subsection{Analysis on 2022 Q1 \& Q2 Data}
To understand how an encoder-decoder LSTM learns the trends of hard-drive sequences, we conduct an analysis of a subset of data to gain insights that can be extended to the larger set. To evaluate the patterns and trends in the data, we selected 2022 Q1 and Q2 data. The primary goal of the analysis was to find connections and relationships between variables, spot patterns in the data, and investigate possible causes and effects of the observed trends. These insights may aid in the identification of areas of interest for further study or  formulation of possible answers to issues within the dataset. 

% We discovered many fundamental problems, such as data scaling, window size configuration, and how to create sequences to conduct useful analysis on the entire dataset. We discovered that encoder-decoder LSTMs can comprehend time series data by learning patterns, and dependencies across time, and retrieving the context encoded in the encoder to predict the RUL sequences.

\subsection{Varying window sizes on 2022 Q1 \& Q2 Data}
\cite{biderectionallstm, hddpaper1, hddpaper2} explore the time-series variation of the S.M.A.R.T. features across a single-sized window. By exploring different window sizes or time steps, an analysis of how the sequence dependence affects the training can be done. We look at different time steps of 5, 10, 15, 20, 25, and 30. 

To make it robust and analyze outcomes better, we run the model with cross-validation evaluation metrics. Our metrics to find the optimal time-step was looking at how each cross-fold varies with training RMSE. In Fig. \ref{fig:window}, window size 25 is an effective time-step that provides variability across multiple training sessions along with a large enough window to predict remaining useful life. Time-steps of 5 or 10 render a very small window to predict whilst 30 is too extreme as far as the variance of training information is concerned.

\subsection{Different Encoder-Decoder LSTM models}

In this subsection, we investigate the impact of different configurations of Long Short-Term Memory (LSTM) units on the performance of the encoder-decoder models. Specifically, we explore six different configurations described in Table \ref{table:configurations}, varying the number of LSTM units in the Encoder and Decoder layers, as well as the number of layers in the Encoder and Decoder. These configurations include 50 LSTM units in both Encoder and Decoder, 100 LSTM units in both, and 200 LSTM units in both. This configuration is a shallower network with just one layer of each kind. Additionally, we investigate the impact of using a single-layer Encoder and multi-layer Decoder, a multi-layer Encoder and single-layer Decoder, and a multi-layer Encoder and multi-layer Decoder. We hypothesize that increasing the number of LSTM units in the Encoder and Decoder layers will improve the model's ability to capture complex relationships between input and output sequences.  To evaluate the performance of these configurations, we ran these models on the Seagate model ST4000DM000. On investigation, we noticed the RMSE values to be lower for the multi-layer configurations but that does not translate to accurate predictions as depicted in Fig. \ref{fig:Ng2}. This points to the conclusion that lower LSTM units with a shallower configuration work best. In the later sections, we confine the experiments to a single-layer encoder and decoder with 50 LSTM units.

% \begin{figure}[htbp]
% \centering
% \begin{subfigure}[b]{0.55\textwidth}
%         \includegraphics[height=6cm]{modelss_1.eps}
%    \caption{}
   
%    \label{fig:Ng1} 
% \end{subfigure}

% \begin{subfigure}[b]{0.55\textwidth}
%    \includegraphics[width=0.88\columnwidth]{RUL_2.eps}
%    \caption{}
%    \label{fig:Ng2}
% \end{subfigure}

% \begin{figure}
%     \centering
%      \includegraphics[width=0.75\columnwidth]{modelss.eps}
%     \caption{Training RMSE evolution for different Encoder-Decoder models: a shallow network is more robust compared to a deep one}
%     \label{fig:Ng1}
% \end{figure}

\begin{table}[htbp]
\centering
\caption{Configurations of the encoder-decoder models}
\label{table:configurations}
\scalebox{0.85}{\begin{tabular}{|c|c|c|c|}
\hline
\textbf{\begin{tabular}[c]{@{}c@{}}Serial \\ Number\end{tabular}} & \textbf{\begin{tabular}[c]{@{}c@{}}Number of LSTM \\ Units per layer\end{tabular}} & \textbf{\begin{tabular}[c]{@{}c@{}}Number of \\ Encoder Layers\end{tabular}} & \textbf{\begin{tabular}[c]{@{}c@{}}Number of \\ Decoder Layers\end{tabular}} \\ 
\hline
\hline
1 & 50 & 1 & 1 \\ \hline
2 & 100 & 1 & 1 \\ \hline
3 & 200 & 1 & 1 \\ \hline
4 & 100 & 1 & 3 \\ \hline
5 & 100 & 3 & 1 \\ \hline
6 & 100 & 2 & 2 \\ \hline
\end{tabular}}
\end{table}

We tabulate RMSE scores for the different configurations presented in Table \ref{table:configurations}. The table below gives an overview of how 50 single-layer encoder and decoder model performs the best with lower RMSE values and with [0, 255] scaling. 

\begin{table}[!htbp]
\centering
\caption{RMSE for the different model configurations}
\label{tab:diffmodelsresult}
\begin{tabular}{|c|ccc|}
\hline
\textbf{Model serial number} & \multicolumn{3}{c|}{\textbf{Log RMSE}}                                                                                    \\ \hline
\textit{\textbf{}}           & \multicolumn{1}{c|}{\textit{\textbf{Train}}} & \multicolumn{1}{c|}{\textit{\textbf{Validation}}} & \textit{\textbf{Test}} \\ \hline
\hline
1                            & \multicolumn{1}{c|}{\textbf{0.83}}                    & \multicolumn{1}{c|}{\textbf{0.75}}                         & \textbf{0.86 }                  \\ \hline
2                            & \multicolumn{1}{c|}{1.42}                    & \multicolumn{1}{c|}{1.41}                         & 1.17                   \\ \hline
3                            & \multicolumn{1}{c|}{1.93}                    & \multicolumn{1}{c|}{1.55}                         & 4.92                   \\ \hline
4                            & \multicolumn{1}{c|}{1.36}                    & \multicolumn{1}{c|}{2.51}                         & 1.49                   \\ \hline
5                            & \multicolumn{1}{c|}{2.05}                    & \multicolumn{1}{c|}{2.04}                         & 2.34                   \\ \hline
6                            & \multicolumn{1}{c|}{1.39}                    & \multicolumn{1}{c|}{2.11}                         & 2.10                   \\ \hline
\end{tabular}
\end{table}

\subsection{Generalizability of the proposed model}
Machine learning models need to be able to perform well on new data that they have not previously encountered, which is why generalizability is an important consideration. A model that possesses strong generalizability can accurately predict outcomes on unseen data. The primary goal of machine learning training is to learn patterns and correlations in the data that can be extended to new data. It is essential to note, however, that obtaining excellent generalizability is not always simple, and can be affected by a number of variables such as the quality and amount of training data, the intricacy of the model, and the underlying distribution at play.

\begin{table}[htbp]
\centering
\caption{Metrics (R2 Score and RMSE) for different Seagate models to depict generalizability}
\label{table:metrics}
\begin{tabular}{|c|c|c|}
\hline
\textbf{Seagate HDD Model} & \textbf{R2 Score} & \textbf{RMSE} 
\\ \hline
\hline
ST4000DM000                & 0.98              & 1.044         \\ \hline
ST12000NM0007              & 0.87              & 2.59          \\ \hline
ST3000DM001                & 0.51              & 5.7           \\ \hline
ST8000NM0055               & 0.6               & 4.52          \\ \hline
ST8000DM002                & 0.68              & 4.08          \\ \hline
ST12000NM0008              & 0.62              & 4.49          \\ \hline
ST31500541AS               & 0.66              & 4.22          \\ \hline
ST31500341AS               & 0.58              & 4.92          \\ \hline
\end{tabular}
\end{table}

In this case, generalizability is challenged on how the proposed network trained on a specific Seagate model behaves on other models of the Seagate family. We train the Encoder-Decoder LSTM model on ST4000DM000 and test the same model on other Seagate models mentioned in Table \ref{table:failures}. We observe that at a certain point, and as the number of failures occurring in the Seagate model decreases, the generalizability suffers. We observe higher RMSE and lower R2 scores for those particular models. These observations are tabulated in Table \ref{table:metrics}. With the two models ST4000DM000 and ST12000NM0007 having failures of 4934 and 2010 respectively, the predicted RUL (seen in Fig. \ref{fig:generalize} is close to the expected values. As we move to other Seagate models with decreasing number of failures in reported logs, the network suffers in its ability to generalize. The predicted RUL is off from the expected RUL, which is expected. Previous research on RUL prediction of HDDs using LSTMs was conducted on relatively short-term data, typically spanning a few months to a few quarters. 

\begin{table}[htbp]
% \centering
\caption{Performance of previous approaches for RUL prediction by various approaches}
\label{tab:results}
\scalebox{0.8}{\begin{tabular}{|c|c|c|c|l|}
\hline
\textbf{Model Type}  & \textbf{Precision} & \textbf{Recall} & \textbf{RMSE} & \textbf{Citation}   \\ \hline
\hline
Random Forest        & 0.95               & 0.67            & -            & Aussel et al. \cite{aussel}       \\ \hline
Attention LSTM       & 0.93               & 0.96            & -            & Wang et al.    \cite{wang}      \\ \hline
Random Forest        & 0.66               & 0.94            & -            & Lu et al. \cite{luetal}           \\ \hline
LSTM                 & 0.66               & 0.88            & -            & Lu et al. \cite{luetal}          \\ \hline
CNN-LSTM             & 0.93               & 0.94            & -            & Lu et al. \cite{luetal}          \\ \hline
TCNN                 & 0.75               & 0.67            & -            & Sun et al. \cite{sun_et_al}         \\ \hline
LSTM                 & 0.98               & 0.98            & -            & Santo et al. \cite{hddpaper4}       \\ \hline
LSTM                 & -                  & -               & 8.15         & Anantharaman et al. \cite{Anantharaman} \\ \hline
Clustered LSTM       & -                  & -               & 2.4          & Basak et al. \cite{BASAK2021101283}       \\ \hline
Bi-LSTM              & -                  & -               & 0.12         & Coursey et al. \cite{biderectionallstm}      \\ \hline
\textbf{Encoder-Decoder LSTM} & -                  & -               & \textbf{0.83}         & \textbf{\textit{Our approach}}        \\ \hline
\end{tabular}}
\end{table}

In contrast, our approach aims to predict RUL for up to 10 years of data, which presents a much more challenging scenario for real-world applications and demonstrates the practicality of our approach over existing methods. We provide a brief overview of results shown by other papers in Table \ref{tab:results}.

\section{Directions for future work}
\label{section:future}
% // Rephrase to make it better

Several topics for future study in hard drive failure prediction using machine learning can be investigated. One avenue of investigation could be to broaden the models' generalizability beyond the Seagate hard drive family. This would entail testing current models against data from other makers and potentially retraining models with new data to improve performance.

Furthermore, investigating the possibility of offline and online training for hard drive failure prediction models could be a fascinating field of study. This would entail looking into how the models work when taught offline and then put on peripheral/edge devices for online training, or when learned using federated learning methods. Another field of investigation could be attention-based LSTMs for hard drive failure forecasts. This would entail examining the efficacy of attention processes in capturing pertinent features and enhancing the models' general performance.

Future studies could also look into the stability of LSTM models when working with threat actor networks.
% This would entail running the models against data containing attack scenarios and evaluating their success in such scenarios. 
There may also be opportunities to leverage recent advancements in explainable AI to provide interpretable insights into the factors driving hard drive failures. By using techniques such as attention visualization or sensitivity analysis, it may be possible to identify specific features or time intervals that are particularly predictive of failure. 
Future research could also look into the use of ensemble networks to forecast hard drive failures from other makers. This would entail training numerous models on data from various makers and then combining their forecasts to improve total performance.
% This could lead to a better understanding of the underlying mechanisms of hard drive failure and more effective maintenance strategies.

% Lastly, 

\section{Conclusion}
This paper employs large-scale predictive analyses using customized feature engineering and sequence learners to develop an encoder-decoder LSTM model that predicts the number of days remaining before a disk potentially fails. The models developed in this work were trained and tested on 10 years of \SMART health data from Backblaze, and the results show that the encoder-decoder LSTM approach is competitive in terms of generalizability over other Seagate family hard drives.

% // Needs work

\section*{Acknowledgment}
The research reported in this publication was supported by the Division of Research and Innovation at San Jos\'e State University under Award Number 23-UGA-08-044. The content is solely the responsibility of the author(s) and does not necessarily represent the official views of San Jos\'e State University.

\bibliographystyle{IEEEtran} % We choose the "plain" reference style
\bibliography{main}

\end{document}